\documentclass[pdflatex,sn-mathphys-num]{sn-jnl}% Math and Physical Sciences Numbered Reference Style
\usepackage[T1]{fontenc}
\usepackage{graphicx}%
\usepackage{multirow}%
\usepackage{amsmath,amssymb,amsfonts}%
\usepackage{amsthm}%
\usepackage{mathrsfs}%
\usepackage[title]{appendix}%
\usepackage{xcolor}%
\usepackage{textcomp}%
\usepackage{manyfoot}%
\usepackage{booktabs}%
\usepackage{algorithm}%
\usepackage{algorithmicx}%
\usepackage{algpseudocode}%
\usepackage{listings}%

\usepackage{amsmath,amssymb,amsfonts}
\usepackage{amsthm}
\usepackage{mathrsfs}
\usepackage{textcomp}
\usepackage{gensymb}

\usepackage{graphicx}
\usepackage{epstopdf}
\usepackage{svg}
\svgsetup{inkscapelatex=false}
\usepackage{placeins}

\usepackage{tikz}
\usetikzlibrary{
    positioning,
    arrows.meta,
    shapes,
    fit,
    matrix,
    calc
}

\usepackage{pgfplots}
\usepackage{pgfplotstable}
\usepgfplotslibrary{groupplots}
\pgfplotsset{compat=1.18}

\usepackage{xcolor}
\usepackage{colortbl}

\usepackage{tabularx}
\usepackage{multirow}
\usepackage{booktabs}
\usepackage{array}
\usepackage{makecell}
\newcolumntype{Y}{>{\centering\arraybackslash}X}

\usepackage{subcaption}

\usepackage{algorithm}
\usepackage{algorithmicx}

\usepackage{enumitem}
\usepackage{placeins}
\usepackage{rotating}
\usepackage{adjustbox}
\usepackage{varwidth}
\usepackage{comment}
\usepackage[normalem]{ulem}
\usepackage[title]{appendix}
\usepackage{listings}

\usepackage{gensymb}
\usepgfplotslibrary{groupplots}
\definecolor{rrtColor}{HTML}{440154}
\definecolor{vfColor}{HTML}{21918C}
\definecolor{svfColor}{HTML}{D95F02}

\usepackage{cleveref}

\usepackage[acronym,nolist]{glossaries}
\makeglossaries

\usepackage[acronym,nolist]{glossaries}
\makeglossaries
\newacronym{usv}{USV}{uncrewed surface vehicle}

\newacronym{mp}{MP}{Microplastics}

\newacronym{swe}{SWE}{Shallow Water Equations}

\newacronym{fe}{FE}{Finite Element}

\newacronym{cfd}{CFD}{Computational Fluid Dynamics}

\newacronym{aoi}{AoI}{Area of Interest}

\newacronym{dwa}{DWA}{Dynamic Window Approach}
\newacronym{tsp}{TSP}{Traveling Salesperson Problem}
\newacronym{cpp}{CPP}{Coverage Path Planning}

\newacronym{ros2}{ROS~2}{Robot Operating System 2}
\newacronym{fem}{FEM}{finite element method}

\newcommand{\yesmark}{\ensuremath{\checkmark}}
\newcommand{\nomark}{\ensuremath{\times}}

\theoremstyle{thmstyleone}%
\theoremstyle{thmstyletwo}%

\theoremstyle{thmstylethree}%

\begin{document}

\title[Article Title]{H-SPAR: Hydrodynamic-aware Simulation for Particle Transport and Autonomous Robots}


\author[1]{\fnm{Navid} \sur{Zarrabi}}

\author[2]{\fnm{Nariman} \sur{Yousefi}}

\author*[3]{\fnm{Sajad} \sur{Saeedi}}
\email{s.saeedi@ucl.ac.uk}

\affil[1]{
\orgdiv{Department of Mechanical, Industrial, and Mechatronics Engineering},
\orgname{Toronto Metropolitan University},
\orgaddress{
\city{Toronto},
\state{Ontario},
\country{Canada}}}

\affil[2]{
\orgdiv{Department of Chemical Engineering},
\orgname{Toronto Metropolitan University},
\orgaddress{
\city{Toronto},
\state{Ontario},
\country{Canada}}}

\affil*[3]{
\orgdiv{Department of Computer Science},
\orgname{University College London},
\orgaddress{
\city{London},
\country{United Kingdom}}}


\abstract{Environmental robotic sampling requires considering the dual influence of water currents on robotic motion and particle transport.  Existing marine robotics simulators generally
model flow, autonomy, and sampling targets separately,
limiting joint evaluation of mission cost and sampling
performance. H-SPAR integrates spatially and temporally
varying velocity fields, Lagrangian particle transport,
probabilistic sampling, and ROS~2/Gazebo-based uncrewed
surface vehicle (USV) autonomy. In this work, shared precomputed flow fields drive particle advection and current-induced forces during closed-loop vehicle execution. Path-planning
experiments show that the existing current-aware planner
SVF-RRT* achieves 69.4\% lower upstream cost than conventional
RRT* at the planning level, but this reduction falls to
41.7\% during execution under time-varying currents,
reflecting temporal flow variation, vehicle motion
constraints, and path deviation omitted during planning.
Coverage experiments show that sweep orientation changes
the particle-sampling rate by up to 22.2\% under the complete
H-SPAR configuration. These findings highlight the importance
of evaluating planning, vehicle execution, particle
transport, and sampling together under consistent
hydrodynamic conditions. The project webpage is available at
\url{https://sites.google.com/view/h-spar},
and the open-source code is available on GitHub at
\url{https://github.com/naviiidz/h-spar-sim}.}




\maketitle

\section{Introduction}\label{sec1}

Marine field robotics has become increasingly important for environmental monitoring, disaster response, and localized pollutant sampling across aquatic ecosystems.
A key challenge in these applications is executing autonomous sampling missions, such as microplastic tracking, oil spill monitoring, or harmful algal bloom characterization, where spatially and temporally varying flow fields actively dictate both system dynamics and target availability.
The varying flow in water currents alter \gls{usv} trajectories, mission duration, and propulsion energy, while simultaneously advecting and redistributing the floating particles or pollutants to be sampled.
Therefore, modeling this dual influence of hydrodynamic flow fields is essential for designing effective autonomous sampling strategies and evaluating true mission performance in real-world marine environments.

Spatially and temporally varying currents create non-convex search spaces where geometrically optimal paths frequently lead to severe trajectory drift, high propulsion expenditure, or actuation saturation. At the same time, because target particles continuously advect with the underlying flow field, the spatial distribution of targets evolves during mission execution. Consequently, evaluating a robot's sampling performance requires accounting for relative fluid-intake velocities and transient particle distributions. Simulating these coupled interactions accurately without incurring the high computational cost of full online \gls{cfd} presents a major bottleneck for current-aware autonomy research.

Prior research has addressed flow-aware marine robotics from complementary but largely separate perspectives. On the algorithmic side, vector-field path planners such as VF-RRT~\cite{ko2013vf} and SVF-RRT~\cite{zhang2023svf} optimize trajectories with respect to prescribed environmental flow fields, while other works explicitly consider uncertainty in time-varying currents~\cite{kularatne2018optimal, lin2023robust}. These algorithms \cite{ko2013vf,zhang2023svf,kularatne2018optimal, lin2023robust} rely on point-mass abstractions while ignoring rigid-body hydrodynamics, focusing primarily on geometric path optimization or energy cost. Likewise, energy-aware coverage approaches often impose simplified environmental assumptions. For example, Zhao et al.~\cite{zhao2024energy} analyze sweep-direction selection under a spatially uniform current direction inside \gls{usv} surveying region of interest. Furthermore, frameworks incorporating temporal updates in water currents discretize hourly forecasts into coarse spatial grids, limiting \gls{usv} mission simulations to static, low-resolution flow-field representations~\cite{fernandes2025smart}.

On the simulation side, established platforms such as DAVE \cite{zhang2022dave} (extending UUV Simulator \cite{manhaes2016uuv}) include aquatic features such as ocean current models (constant, stratified, and tidal), dynamic bathymetry spawning, and physics-based underwater sensors for underwater robots. Similarly, the Virtual RobotX (VRX) simulator~\cite{bingham2019toward} extends Gazebo with wave, wind, buoyancy, and surface vessel dynamics models to support autonomy testing on \glspl{usv}, with potential of using Gazebo plugins for adding water currents on \glspl{usv}. In addition, MOOS-IvP~\cite{benjamin2010nested} has been widely used for \gls{usv} mission planning and multi-vehicle coordination, offering a mature autonomy middleware with integrated simulation capabilities. MOOS-IvP uses additional current toolboxes to accept externally published drift vectors. Photorealistic simulators such as MARUS~\cite{lonvcar2022marus} and UNav-Sim~\cite{amer2023unav} prioritize visual rendering of aquatic environments for perception tasks. MARUS represent basic irrotational current fields, whereas UNav-Sim abstracts water currents into simplified external force disturbances (modeled as constant or sinusoidal functions). None of the above platforms have the ability to ingest fine-grained spatio-temporal velocity fields and couple them with Lagrangian particle tracking and probabilistic intake models (Table~\ref{tab:simulator_comparison}). As a result, researchers cannot jointly evaluate \gls{usv} trajectory execution, energy expenditure, and environmental sampling efficiency within a single closed-loop simulator.

\begin{table}[t]
    \caption{Comparison of marine robotics simulators.
    H-SPAR combines solver-derived flow fields, current-affected
    USV execution, Lagrangian particle transport, and
    probabilistic particle sampling.}
    \label{tab:simulator_comparison}

    \small
    \renewcommand{\arraystretch}{1}

    \begin{tabular*}{\textwidth}{@{\extracolsep\fill}lcccccc}
        \toprule
        \textbf{Feature}
        & \rotatebox{70}{\textbf{DAVE}~\cite{zhang2022dave}}
        & \rotatebox{70}{\textbf{VRX}~\cite{bingham2019toward}}
        & \rotatebox{70}{\textbf{MOOS-IvP}~\cite{benjamin2010nested}}
        & \rotatebox{70}{\textbf{UNav-Sim}~\cite{amer2023unav}}
        & \rotatebox{70}{\textbf{MARUS}~\cite{lonvcar2022marus}}
        & \rotatebox{70}{\textbf{H-SPAR}} \\
        \midrule

        Current/disturbance modeling
            & \yesmark & \yesmark & \yesmark & \yesmark & \yesmark & \yesmark \\

        Solver-derived flow field
            & \nomark & \nomark & \nomark & \nomark & \nomark & \yesmark \\

        Current-to-vehicle coupling
            & \yesmark & \yesmark & \nomark & \yesmark & \yesmark & \yesmark \\

        Dynamic vehicle execution
            & \yesmark & \yesmark & \yesmark & \yesmark & \yesmark & \yesmark \\

        Photorealistic rendering
            & \nomark & \nomark & \nomark & \yesmark & \yesmark & \nomark \\

        Lagrangian particle transport
            & \nomark & \nomark & \nomark & \nomark & \nomark & \yesmark \\

        Probabilistic particle sampling
            & \nomark & \nomark & \nomark & \nomark & \nomark & \yesmark \\

        \botrule
    \end{tabular*}

    \footnotetext{%
    \yesmark~indicates explicitly supported functionality;
    \nomark~indicates functionality not modeled natively.
    Hydrodynamic coupling refers to calculations derived from
    local flow velocities.}

\end{table}

In marine environments, suspended particles and \glspl{usv} are influenced by the same underlying flow field. Evaluating particle-sampling missions therefore requires the joint simulation of flow-driven particle transport and current-affected \gls{usv} motion. Although recent studies have investigated robotic platforms for in situ monitoring of particles and plumes~\cite{faltynkova2024use, cho2025capillary}, existing marine robotics simulators provide no support for jointly modeling hydrodynamic particle transport, flow-dependent robot execution, and sampling-oriented mission planning. Such joint simulation can be performed at different levels of physical fidelity. Particle-based fluid methods can represent detailed fluid behavior and particle-scale interactions; however, enforcing physical properties such as incompressibility introduces substantial computational demands~\cite{macklin2013position}. Explicitly resolving these details for large particle populations is therefore not well suited to mission-scale marine simulations, where the primary objective is to capture particle transport and spatial distribution.

To address these limitations, we propose {H-SPAR}, an open-source and hydrodynamic-aware simulation framework for particle transport and autonomous marine robotics. 
As shown in Fig.~\ref{fig::va}, {H-SPAR} uses a shared fine-grained velocity-field representation, the framework simultaneously models current-induced wrenches acting on the \gls{usv} and the advection of surrounding target particles.

\usetikzlibrary{
    arrows.meta,
    positioning,
    calc,
    fit,
    backgrounds
}
\begin{figure}[t]
    \centering
    \begin{tikzpicture}
        % Load the image node
        \node[inner sep=0pt, anchor=south west] (img) at (0,0) {
            \includegraphics[width=0.85\linewidth]{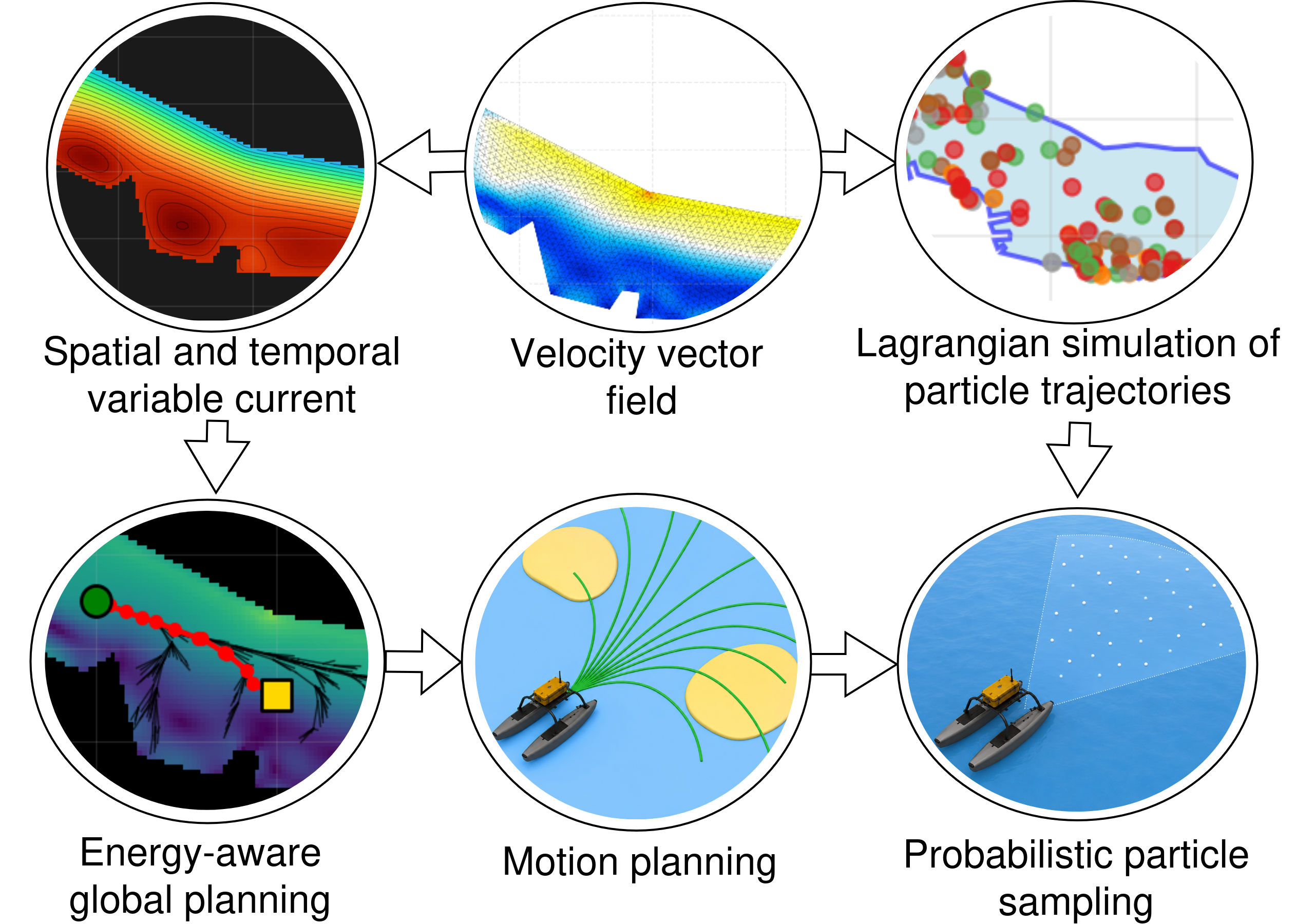}
        };
        
        % Overlay labels over the image using percentage-based relative coordinates (0 to 1)
        % Adjust positioning (x, y) if needed:

        \node[anchor=north west, fill=white, fill opacity=0.8, text opacity=1, inner sep=2pt, rounded corners=1pt] 
            at ([shift={(3.8,-0.02)}]img.north west) {\small (a)};
            %at ([shift={(2.5,-0.02)}]img.north west) {\bfseries\small (a)};

        \node[anchor=north west, fill=white, fill opacity=0.8, text opacity=1, inner sep=2pt, rounded corners=1pt] 
            at ([shift={(0.02,-0.02)}]img.north west) {\small (c)};
            
        \node[anchor=north west, fill=white, fill opacity=0.8, text opacity=1, inner sep=2pt, rounded corners=1pt] 
            at ([shift={(7.5,-0.02)}]img.north west) {\small (b)};

        \node[anchor=north west, fill=white, fill opacity=0.8, text opacity=1, inner sep=2pt, rounded corners=1pt] 
            at ([shift={(3.8,-4.2)}]img.north west) {\small (e)};

        \node[anchor=north west, fill=white, fill opacity=0.8, text opacity=1, inner sep=2pt, rounded corners=1pt] 
            at ([shift={(0.02,-4.2)}]img.north west) {\small (d)};
            
        \node[anchor=north west, fill=white, fill opacity=0.8, text opacity=1, inner sep=2pt, rounded corners=1pt] 
            at ([shift={(7.7,-4.2)}]img.north west) {\small (f)};            
    \end{tikzpicture}
\caption{H-SPAR uses (a) fine-grained velocity vector fields to simultaneously 
(b) simulate the advective transport of particles and (c) model spatially and temporally varying currents. The variable currents are used to  
(d) calculate hydrodynamic drag forces at the \gls{usv} reference frame. 
By integrating Lagrangian particle transport with hydrodynamic-aware \gls{usv} simulation (e-f) H-SPAR enables the evaluation of particle-sampling missions under marine flow conditions.
}
    \label{fig::va}
\end{figure}
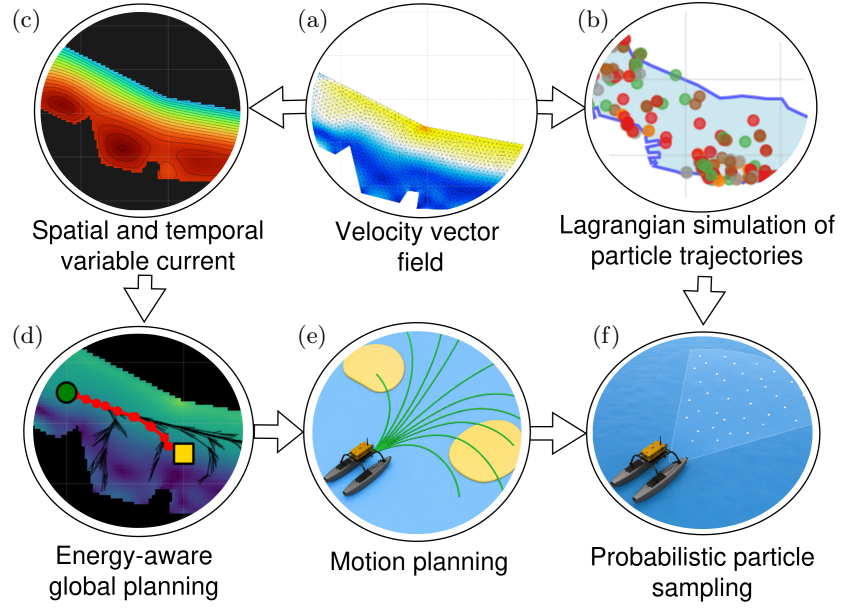

Table~\ref{tab:simulator_comparison} highlights the key advantages of H-SPAR compared with existing works. 
%
%H-SPAR integrates physics-based environmental flow simulation, Lagrangian particle transport, \gls{ros2} middleware, and Gazebo vessel dynamics within a unified closed-loop environment (Fig.~\ref{fig:front-back-structure}). 
The primary contributions of this work are as follows:
\begin{itemize}
    \item {Unified System Architecture:} An open-source \gls{ros2}/Gazebo framework that bridges environmental flow fields with closed-loop \gls{usv} control pipelines, enabling synchronized simulation of vehicle dynamics and environmental transport.
    \item {Lagrangian and Probabilistic Sampling Module:} A physics-informed particle tracking module incorporating a relative-velocity capture function to model \gls{usv}-particle intake interactions under hydrodynamic flow.
    \item {Planning vs. Execution Evaluation Pipeline:} A benchmarking pipeline evaluating path planners ($RRT^*$, $VF\text{-}RRT^*$, $SVF\text{-}RRT^*$) across both planning-stage costs and realized closed-loop execution metrics (trajectory drift, upstream cost, execution time).
    \item {Hydrodynamic Coverage Benchmarking:} A mission-level evaluation demonstrating how sweep orientation in boustrophedon coverage impacts particle capture rate, mission duration, and energy expenditure under realistic flow conditions.
\end{itemize}    
%\end{addblock}

The H-SPAR simulator, including experiment scripts, datasets, and fine-grain velocity field generators, will be open-source and will be available via the project page: \url{https://sites.google.com/view/h-spar}. For the rest of the paper, Sec.~\ref{methodology} introduces the structure of the simulator, detailing its front-end and back-end components. Sec.~\ref{experiments} describes how this integration enables an environment-aware tool for robotic mission planning, and Sec.~\ref{conclusion} concludes the paper.

\section{Methodology}
\label{methodology}
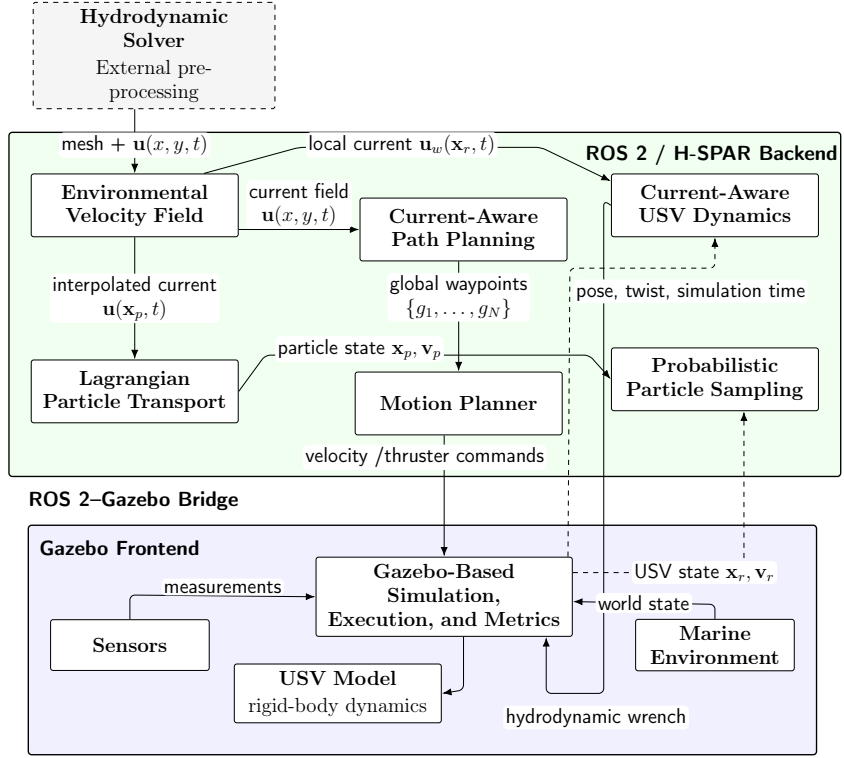
\begin{figure*}[b]
\centering
\resizebox{0.85\textwidth}{!}{%
\begin{tikzpicture}[
    font=\sffamily,
    component/.style={
        draw,
        rounded corners=1.5pt,
        line width=0.6pt,
        minimum width=4cm,
        minimum height=1.30cm,
        text width=4cm,
        align=center,
        inner sep=4pt,
        fill=white,
        font=\large
    },
    smallcomponent/.style={
        draw,
        rounded corners=1.5pt,
        line width=0.6pt,
        minimum width=3cm,
        minimum height=1.05cm,
        text width=3cm,
        align=center,
        inner sep=4pt,
        fill=white,
        font=\large
    },
        regularcomponent/.style={
        draw,
        rounded corners=1.5pt,
        line width=0.6pt,
        minimum width=4cm,
        minimum height=0.85cm,
        text width=4cm,
        align=center,
        inner sep=4pt,
        fill=white,
        font=\large
    },
    external/.style={
        draw,
        dashed,
        rounded corners=1.5pt,
        line width=0.6pt,
        minimum width=5cm,
        minimum height=1.0cm,
        text width=4cm,
        align=center,
        inner sep=4pt,
        fill=gray!8,
        font=\large
    },
    rospanel/.style={
        draw,
        rounded corners=2pt,
        line width=0.8pt,
        fill=green!6,
        inner xsep=0.45cm,
        inner ysep=0.85cm
    },
    gzpanel/.style={
        draw,
        rounded corners=2pt,
        line width=0.8pt,
        fill=blue!6,
        inner xsep=0.45cm,
        inner ysep=0.65cm
    },
    dataarrow/.style={
        -{Latex[length=2mm]},
        line width=0.6pt,
        rounded corners=3pt
    },
    returnarrow/.style={
        -{Latex[length=2mm]},
        line width=0.6pt,
        dashed,
        rounded corners=3pt
    },
    datalabel/.style={
        font=\sffamily\large,
        align=center,
        fill=white,
        inner sep=1.5pt
    },
    subsystemtitle/.style={
        font=\sffamily\large\bfseries,
        anchor=north west
    }
]

% ============================================================
% EXTERNAL HYDRODYNAMIC SOLVER
% ============================================================

\node[external] (solver) at (-6.8,0.2) {
    \textbf{Hydrodynamic Solver}\\[2pt]
    {\large External preprocessing}
};

% ============================================================
% ROS 2 / H-SPAR -- TOP ROW
% ============================================================

\node[component] (field) at (-7.2,-2.9) {
    \textbf{Environmental}\\
    \textbf{Velocity Field}
};

\node[component] (planner) at (-0.4,-3.4) {
    \textbf{Current-Aware}\\
    \textbf{Path Planning}
};

\node[component] (dynamics) at (4.8,-2.9) {
    \textbf{Current-Aware}\\
    \textbf{USV Dynamics}
};

% ============================================================
% ROS 2 / H-SPAR -- BOTTOM ROW
% ============================================================

\node[component] (particle) at (-7.2,-6.75) {
    \textbf{Lagrangian}\\
    \textbf{Particle Transport}
};

\node[component] (controller) at (-0.5,-7) {
    \textbf{Motion Planner}\\[2pt]
};

\node[component] (sampling) at (4.8,-6.5) {
    \textbf{Probabilistic}\\
    \textbf{Particle Sampling}
};

% ============================================================
% ROS 2 BACKGROUND
% ============================================================

\begin{scope}[on background layer]
\node[
    rospanel,
    fit=(field)(planner)(dynamics)(particle)(controller)(sampling)
] (rosbox) {};
\end{scope}

\node[subsystemtitle]
at ($(rosbox.north)+(3.2,-0.15)$)
{ROS 2 / H-SPAR Backend};

% ============================================================
% GAZEBO FRONTEND
% ============================================================

\node[
    component,
    minimum width=5.3cm,
    text width=5.0cm,
    minimum height=1.55cm
] (gazebo) at (-0.8,-11) {
    \textbf{Gazebo-Based Simulation,}\\
    \textbf{Execution, and Metrics}
};

\node[smallcomponent] (sensors) at (-7.3,-12) {
    \textbf{Sensors}\\[2pt]

};

\node[smallcomponent] (environment) at (4.8,-12) {
    \textbf{Marine}\\
    \textbf{Environment}
};

\node[regularcomponent] (usv) at (-3,-13) {
    \textbf{USV Model}\\[2pt]
    rigid-body dynamics
};

% ============================================================
% GAZEBO BACKGROUND
% ============================================================

\begin{scope}[on background layer]
\node[
    gzpanel,
    fit=(sensors)(gazebo)(environment)(usv)
] (gzbox) {};
\end{scope}

\node[subsystemtitle]
at ($(gzbox.north west)+(0.12,-0.2)$)
{Gazebo Frontend};

% ============================================================
% EXTERNAL INPUT
% ============================================================

\draw[dataarrow]
    ($(solver.south)+(-0.4,0)$)
    --
    node[datalabel, yshift=-0.8mm]
    {mesh + $\mathbf{u}(x,y,t)$}
    (field.north);

% ============================================================
% VELOCITY FIELD -> PATH PLANNING
% ============================================================

\draw[dataarrow]
    ($(field.east)+(0,-0.5)$)
    --
    node[datalabel, above]
    {current field\\$\mathbf{u}(x,y,t)$}
    (planner.west);

% ============================================================
% VELOCITY FIELD -> USV DYNAMICS
%
% Route this above the top row so it does not pass through
% the path-planning block.
% ============================================================

\coordinate (flowlaneA) at (-3.8,-1.70);
\coordinate (flowlaneB) at ( 1.2,-1.70);
\coordinate (flowlaneC) at ( 2,-2);

\draw[dataarrow]
    ($(field.north east)+(-0.72,0)$)
    -- (flowlaneA)
    -- (flowlaneB)
    --
    node[datalabel, yshift=4mm, xshift=-36mm]
    {local current $\mathbf{u}_{w}(\mathbf{x}_r,t)$}
    ($(dynamics.west)+(0,0.5)$);

% ============================================================
% VELOCITY FIELD -> PARTICLE TRANSPORT
% ============================================================

\draw[dataarrow]
    (field.south)
    --
    node[datalabel ]
    {interpolated current\\
     $\mathbf{u}(\mathbf{x}_p,t)$}
    (particle.north);

% ============================================================
% GLOBAL PATH -> LOCAL PLANNER
% ============================================================

\draw[dataarrow]
    ($(planner.south)+(-0.1,0)$)
    --
    node[datalabel, yshift=0.4cm]
    {global waypoints\\
     $\{g_1,\ldots,g_N\}$}
    (controller.north);

% ============================================================
% PARTICLE TRANSPORT -> SAMPLING
%
% Use a lower routing lane so the arrow does not pass through
% the local planner/controller.
% ============================================================

\coordinate (particleLaneA) at (-4.5,-6);
\coordinate (particleLaneB) at ( 2.2,-6);

\draw[dataarrow]
    (particle.east)
    -- (particleLaneA)
    --
    node[datalabel, below, yshift=4mm, xshift=-14mm]
    {particle state $\mathbf{x}_p,\mathbf{v}_p$}
    (particleLaneB)
    --
    (sampling.west);
% ============================================================
% SAMPLING MODEL EQUATION
% ============================================================

% ============================================================
% ROS 2--GAZEBO BRIDGE LABEL
% ============================================================

\node[
    font=\sffamily\large\bfseries,
    fill=white,
    inner sep=0.5pt,
    xshift=-60mm
]
at ($(rosbox.south)!0.50!(gzbox.north)$)
{ROS 2--Gazebo Bridge};

% ============================================================
% LOCAL CONTROLLER -> GAZEBO
%
% This can now go straight down because the controller is
% centered directly above Gazebo.
% ============================================================

\draw[dataarrow]
    ($(controller.south)+(-0.3,0)$)
    --
    node[datalabel, yshift=8mm, xshift=-4mm]
    {velocity /thruster commands}
    (gazebo.north);

% ============================================================
% DYNAMICS -> GAZEBO
% HYDRODYNAMIC WRENCH
%
% Dedicated right-side lane.
% ============================================================

\coordinate (wrenchA) at (2.5,-2.8);
\coordinate (wrenchB) at (2.5,-13);

\draw[dataarrow]
    (dynamics.west)
    -- (wrenchA)
    -- (wrenchB)
    -|
    node[datalabel, pos=0.50, right, yshift=-5mm, xshift=-9mm]
    {hydrodynamic wrench}
    ($(gazebo.south east)+(-0.55,0)$);

% ============================================================
% GAZEBO -> DYNAMICS
% POSE, TWIST, SIMULATION TIME
%
% Separate outer lane prevents overlap with wrench arrow.
% ============================================================

\coordinate (stateA) at (1.8,-4.2);
\coordinate (stateB) at (4.8,-4.2);

\draw[returnarrow]
    ($(gazebo.north east)+(-0.1,0)$)
    -- (stateA)
    -- (stateB)
    --
    node[datalabel, yshift=-8mm, xshift=-5mm]
    {pose, twist, simulation time}
    (dynamics.south);

% ============================================================
% GAZEBO -> PARTICLE SAMPLING
%
% USV state required for relative particle/robot velocity.
% ============================================================

\draw[returnarrow]
    ($(gazebo.east)+(0,0.5)$)
    -|
    node[datalabel, pos=0.3, xshift=6mm]
    {USV state $\mathbf{x}_r,\mathbf{v}_r$}
    ($(sampling.south)+(0.6,0)$);

% ============================================================
% INTERNAL GAZEBO CONNECTIONS
% ============================================================

\draw[dataarrow]
    (sensors.north)
    -- ++(0,0.5)
    --
    node[datalabel, above]
    {measurements}
    (gazebo.west);

\draw[dataarrow]
    (environment.north)
    -- ++(0,0.3)
    --
    node[datalabel]
    {world state}
    ($(gazebo.east)+(0,-0.1)$);

\draw[dataarrow]
    ($(gazebo.south)+(0.4,0)$)
    -- ++(0,-1.1)
    --
    (usv.east);

\end{tikzpicture}
}

\caption{Overview of the H-SPAR simulation framework, showing the integration between
ROS~2-based planning and hydrodynamic modules and the Gazebo-based marine
simulation environment. ROS~2 hosts current-aware path and motion planning,
hydrodynamic flow modeling, and Lagrangian particle transport, while Gazebo provides the simulated aquatic environment, virtual sensors, and \gls{usv} dynamic models. Sensor measurements are transmitted from Gazebo to ROS~2, while ROS~2 provides thruster commands and hydrodynamic forces to the simulated vehicles, enabling closed-loop flow-aware navigation and particle monitoring.}
\label{fig:front-back-structure}

\end{figure*}

\textcolor{black}{H-SPAR integrates a hydrodynamic-aware backend implemented in ROS~2 with a physics-aware frontend implemented in Gazebo. Fig.~\ref{fig:front-back-structure} shows an overview of the method. Before runtime, spatially and temporally varying environmental velocity fields are generated using Firedrake~\cite{rathgeber2016firedrake} and provided to H-SPAR as input. During simulation, the backend uses precomputed fields to update particle locations and compute the current-induced drag force acting on the \gls{usv}. The frontend applies the computed force to the \gls{usv} model while simulating the vehicle dynamics, environmental physics, and onboard sensors.}

\textcolor{black}{The modular separation between environmental flow modeling and Gazebo-based vehicle dynamics also allows additional environmental disturbances to be incorporated independently of the current field. In particular, H-SPAR inherits the wave and sea-state simulation capabilities of the VRX simulator~\cite{saldarriaga2025open} used in the Gazebo frontend. Configurable wave conditions can therefore be introduced through parameters such as wave height, period, and propagation direction, while H-SPAR independently provides the spatially and temporally varying current field.}

\textcolor{black}{The methodology first describes the representation and integration of the environmental velocity fields (Sec.~\ref{sec::hydro-flow-modeling}), followed by current-aware path planning (Sec.~\ref{sec::path-planning}) and the computation of current-induced forces acting on the \gls{usv} (Sec.~\ref{sec::motion-planner}). The planned missions are then executed in Gazebo, where vehicle behavior and mission performance are assessed using execution-level metrics (Sec.~\ref{sec::ros-gz}). The same velocity fields drive the Lagrangian particle-transport model (Sec.~\ref{sec::lag-particle-trans}), while interactions between the \gls{usv} and nearby particles are represented using a probabilistic particle-sampling model (Sec.~\ref{sec::sampling}).}

In this paper, we adopt these notations:
lowercase italic (\(x, y, z, \eta, b\)) for scalars, 
lowercase bold (\(\mathbf{u}, \mathbf{v}, \mathbf{r}\)) for vectors, and 
uppercase bold (\(\mathbf{M}, \mathbf{C}, \mathbf{D}\)) for matrices or operators.

\subsection{Environmental Velocity Field}
\label{sec::hydro-flow-modeling}

\textcolor{black}{H-SPAR represents the hydrodynamic environment using fine-grained velocity fields that describe the local water motion. These fields can be obtained from different sources, including physics-based hydrodynamic solvers~\cite{matsson2022introduction}, analytical flow models~\cite{zhang2023svf}, and data-driven predictors~\cite{pfaff2020learning}, depending on the target environment and the desired trade-off between computational cost and fidelity. H-SPAR is therefore independent of the method used to generate the velocity fields, provided that the resulting data contain the spatial and temporal information required by the simulation.}

\textcolor{black}{In this work, the computational domain is discretized using an unstructured triangular mesh generated with Gmsh~\cite{geuzaine2009gmsh}. The unstructured representation accommodates irregular shorelines, concave regions, and obstacle boundaries~\cite{hervouet2007hydrodynamics}, while allowing local mesh refinement in regions with complex geometry or strong flow gradients~\cite{wallwork2020goal}. The resulting velocity fields are exported and subsequently used by H-SPAR for particle advection, current-aware path evaluation, and \gls{usv} mission simulation.}

\textcolor{black}{To balance hydrodynamic fidelity and computational scalability, the velocity fields used in this study are generated using the depth-averaged \gls{swe}~\cite{karna2018thetis}. This formulation captures large-scale horizontal flow structures while remaining computationally tractable for mission-level robotic simulation. Because the equations describe depth-averaged horizontal motion, they do not resolve vertical velocity profiles, stratification, three-dimensional turbulence, or near-body fluid interactions. These assumptions are appropriate for the surface-level transport scenarios considered here, in which particle advection and \gls{usv} motion are primarily influenced by large-scale horizontal currents.}

% \begin{figure*}[t]
%     \centering
%     \includegraphics[width=1\linewidth]{figures/H-SPAR-2.png}
%     \caption{Overview of the H-SPAR simulation framework, showing the integration between Gazebo-based marine simulation assets and ROS2-based planning and hydrodynamic modules. Gazebo provides virtual sensors, USV models, and aquatic environments, while ROS2 hosts energy-aware planners, motion planning, hydrodynamic modeling, and particle simulation. Bidirectional communication enables closed-loop testing of autonomous surface vehicle navigation and microplastic monitoring under flow-aware environmental conditions.}
%     \label{fig:front-back-structure}
% \end{figure*}

\subsection{Current-Aware Path Planning}
\label{sec::path-planning}

Water currents can substantially affect the hydrodynamic difficulty, feasibility, and duration of marine robotic missions. A path that is geometrically short may require prolonged upstream motion and therefore be difficult to execute efficiently. H-SPAR represents this effect using an upstream-cost metric~\cite{zhang2023svf} that depends on the motion of the \gls{usv} relative to the local current field.

H-SPAR implements and evaluates the geometry-based RRT* planner and two current-aware variants, VF-RRT*~\cite{ko2013vf} and SVF-RRT*~\cite{zhang2023svf}. RRT* considers path length and obstacle avoidance, whereas VF-RRT* and SVF-RRT* additionally query the environmental velocity field along candidate path segments. This allows the planners to favor trajectories that are more aligned with the local flow and avoid regions with strong opposing currents.

All planners use the same static navigation costmap to represent traversable water, land boundaries, obstacles, and no-go regions. Each planner outputs a sequence of waypoints, which is passed to the \gls{ros2} execution node. This separation enables H-SPAR to compare the cost of the planned path with the behavior of the trajectory after execution under vehicle constraints and current-induced disturbances.

To quantify the hydrodynamic difficulty of both planned and executed trajectories, H-SPAR uses a discrete upstream-cost metric adapted from Eq.~(1) of SVF-RRT*~\cite{zhang2023svf}. At the planning level, the metric is evaluated along the generated path using the local velocity field.
% Given a path $P=(\mathbf{p}_0,\mathbf{p}_1,\ldots,\mathbf{p}_{N_s})$, each segment is 
% defined as $\mathbf{s}_i=\mathbf{p}_{i+1}-\mathbf{p}_i$, with length 
% $\Delta s_i=\|\mathbf{s}_i\|$ and unit tangent 
% $\mathbf{t}_i=\mathbf{s}_i/\Delta s_i$. The local current vector 
% $\mathbf{f}_i$ is sampled at the segment midpoint. The upstream cost is then 
% computed as:
% \begin{equation}
%     \color{black}
%     \hat{c}(P)=
%     \sum_{i=0}^{N_s-1}
%     \left(
%     \|\mathbf{f}_i\|-\mathbf{f}_i^\top \mathbf{t}_i
%     \right)\Delta s_i .
%     \label{eq:upstream_cost}
% \end{equation}
\subsection{Current-Aware \gls{usv} Dynamics}\label{sec::motion-planner}
H-SPAR estimates current-induced drag based on the spatial and temporal state of each \gls{usv}. At each simulation time step, the framework queries the vehicle position and simulation timestamp. It then retrieves the corresponding local current velocity and computes the resulting drag force and moment.

Accurately modeling hydrodynamic drag is essential for simulating \gls{usv} dynamics in current-influenced environments. We adopt a quadratic damping formulation commonly used in marine robotics to represent surge, sway, and yaw effects~\cite{bingham2019toward}. This formulation is suitable for control-oriented \gls{usv} models, where hydrodynamic coefficients and mass properties are often uncertain and difficult to characterize analytically~\cite{klinger2016control}.

In our \textcolor{black}{design}, the \gls{usv} features two hulls in a catamaran layout, providing a wide stance and reduced drag. For each hull $i$, the relative flow velocity is obtained as the difference between the current velocity and the local hull velocity, including rotational contributions~\cite{fossen2011handbook}:
\begin{equation}
\mathbf{u}_{\text{rel},i} = 
\left(
\begin{bmatrix}
u_i \\ v_i
\end{bmatrix}
+
\begin{bmatrix}
-\omega_i r_{y,i}\\
\omega_i r_{x,i}
\end{bmatrix}
\right)-
\mathbf{u}_{w}^{}~, 
\end{equation}
\noindent where $\mathbf{u}_w^{\text{}}$ is the current velocity vector expressed in the hull body frame, $(u_i,v_i)$ are body-frame linear velocities, $\omega_i$ is the yaw rate, and $\mathbf{r}_i=(r_{x,i},r_{y,i})$ is the drag application point relative to the vessel origin. Hydrodynamic drag in the surge and sway directions is modeled as~\cite{fossen2011handbook}:
\begin{equation}
\mathbf{F}_i =
-\tfrac{1}{2}\rho
\begin{bmatrix}
C_{D,x,i} A_{x,i} \, |u_{\text{rel},x,i}| u_{\text{rel},x,i} \\
C_{D,y,i} A_{y,i} \, |u_{\text{rel},y,i}| u_{\text{rel},y,i}
\end{bmatrix},
\end{equation}

\noindent where $\rho$ is the water density, 
$C_{D,x,i}$ and $C_{D,y,i}$ are the empirical drag coefficients for hull $i$ in surge and sway, 
and $A_{x,i}$ and $A_{y,i}$ denote the corresponding effective projected areas. 
The yaw torque generated by drag on hull $i$ is:
\begin{equation}
\tau_{z,i} = r_{x,i} F_{y,i} - r_{y,i} F_{x,i}
= \big(\mathbf{r}_i \times \mathbf{F}_i\big)_z,
\end{equation}
\noindent where $\mathbf{F}_i=(F_{x,i},F_{y,i})^\top$ is the corresponding drag force vector. To approximate the motion of a marine vehicle in the presence of ocean currents, we modify the 6-DOF model expressed in~\cite{bingham2019toward}:
\begin{align}
\mathbf{M}_{RB}\dot{\mathbf{u}} 
&+ \mathbf{C}_{RB}(\mathbf{u})\,\mathbf{u}
+ \mathbf{M}_A \dot{\mathbf{u}}_r 
+ \mathbf{C}_A(\mathbf{u}_r)\,\mathbf{u}_r \notag \\
&+ \mathbf{D}(\mathbf{u}_r)\,\mathbf{u}_r 
+ \mathbf{g}(\boldsymbol{\eta}) \notag \\
&= \boldsymbol{\tau}_{\text{propulsion}} 
+ \boldsymbol{\tau}_{\text{wind}}
+ \boldsymbol{\tau}_{\text{waves}} 
+ \boldsymbol{\tau}_{\text{current}}~.
\label{eq:fossen6dof}
\end{align}

\noindent

Here, the generalized position vector is
$\eta$ ,
which contains the earth-fixed Cartesian position and the Euler orientation angles corresponding to roll, pitch, and yaw. The generalized velocity vector is $\boldsymbol{u}$,
comprising of the body-fixed linear velocities in surge, sway, and heave, and the angular velocities about roll, pitch, and yaw. The current-relative velocity is $\boldsymbol{u}_r$, which ensures that the hydrodynamic forces are evaluated relative to the moving fluid.

Eq.~\eqref{eq:fossen6dof} includes the rigid-body~(RB) inertia and Coriolis forces due to the \gls{usv}’s mass and geometry, and added mass inertia and Coriolis forces from the surrounding fluid. It also represents linear and nonlinear hydrodynamic damping. The vector \(\boldsymbol{g(\eta)}\) models the hydrostatic restoring forces due to gravity and buoyancy.
The generalized inputs
\(\boldsymbol{\tau}_{\text{propulsion}},\;\boldsymbol{\tau}_{\text{wind}},\;\boldsymbol{\tau}_{\text{waves}},\;\boldsymbol{\tau}_{\text{current}}\)
denote, respectively, the propulsion forces from thrusters, aerodynamic loads from wind, wave excitation forces and moments, and the external loads due to spatially varying currents. In this work, a current-relative velocity is computed by querying the current field at the vehicle’s global position \((x,y,z)\).

\subsection{Gazebo-Based Simulation, Execution, and Metrics}
\label{sec::ros-gz}

{\color{black}Gazebo provides the physics-based simulation frontend of H-SPAR. The Gazebo ecosystem includes reusable marine environments, \gls{usv} models, sensors, obstacles, and plugins for buoyancy and hydrodynamic effects. It also simulates rigid-body dynamics, propulsion, collision handling, and onboard sensor measurements.

During simulation, the pose and velocity of the \gls{usv} are continuously transmitted through the ROS 2-Gazebo bridge. An H-SPAR ROS 2 node uses the vehicle state and simulation timestamp to query the local current field. \textcolor{black}{Before simulation, a KD-tree~\cite{kd-tree}, implemented using
SciPy~\cite{virtanen2020scipy},  is constructed from the fixed mesh coordinates to enable efficient spatial indexing. Since the mesh geometry remains unchanged throughout the simulation, the KD-tree is generated only once during initialization. At runtime, the vehicle position is queried against the KD-tree to obtain the index of the nearest mesh point, and the corresponding velocity vector is retrieved using the simulation timestamp and spatial index.} The corresponding drag force and moment are computed using the formulation described in Sec.~\ref{sec::motion-planner}. The resulting hydrodynamic wrench is then applied to the \gls{usv} model through the bridge.}
{\color{black}
For low-level vehicle actuation, H-SPAR adopts the thrust-mapping and model-based control approach of Saldarriaga~et~al.~\cite{saldarriaga2025open}. This method converts linear and angular velocity references into individual commands for thrusters. H-SPAR uses this existing controller as part of the execution pipeline and does not introduce a new actuator-control formulation.

During each mission, H-SPAR records planning-level and execution-level data to evaluate geometric path cost, hydrodynamic difficulty, and trajectory feasibility. At the planning level, the planned path length is computed as the total Euclidean distance between consecutive waypoints generated by the path planner. The planned upstream cost is evaluated along the same waypoint sequence using the current-aware cost formulation described in Sec.~\ref{sec::path-planning}.

At the execution level, H-SPAR records the actual trajectory length, mission duration, local current velocity, applied hydrodynamic wrench, and executed upstream cost. The executed upstream cost is evaluated along the recorded trajectory using the same formulation applied at the planning level. This enables direct comparison between the predicted cost of a planned path and its realized performance during simulation.

To quantify deviation from the planned path, each executed position is compared with the closest segment of the planned polyline. The point-wise path deviation is defined as
\begin{equation}
    d_i =
    \min_j
    \operatorname{dist}
    \left(
    \mathbf{p}_i,\,
    \overline{\mathbf{g}_j\mathbf{g}_{j+1}}
    \right),
    \label{eq:pointwise_path_deviation}
\end{equation}
where $\mathbf{p}_i$ is the $i$-th recorded vehicle position and
$\overline{\mathbf{g}_j\mathbf{g}_{j+1}}$ denotes the line segment between consecutive planned waypoints. The mean path deviation is then computed as
\begin{equation}
    D =
    \frac{1}{N}
    \sum_{i=1}^{N}
    d_i,
    \label{eq:mean_path_deviation}
\end{equation}
where $N$ is the number of recorded trajectory points.

Together, these metrics allow planned and executed trajectories to be compared in terms of path length, upstream cost, mission duration, and tracking deviation.}

\subsection{Lagrangian Particle Transport}
\label{sec::lag-particle-trans}

The fine-grained Eulerian velocity fields are interpolated at each particle location to update the motion of Lagrangian particles in the water. Particle transport accounts for both advection by the local flow velocity and diffusion~\cite{pilechi2022numerical}.
\begin{equation}
\frac{d\boldsymbol{x}}{dt} = \boldsymbol{u}(\boldsymbol{x},t) + \boldsymbol{R}(t),
\end{equation}

\noindent with $\boldsymbol{x}(t)$ denoting particle position, $\boldsymbol{u}(x,t)$ the interpolated velocity field, and $\boldsymbol{R}(t)$ an optional stochastic term for subgrid-scale turbulent diffusion. This model prevents numerical diffusion, which smooths sharp concentration gradients and obscures fine-scale structures common in Eulerian methods~\cite{wagner2019can}.\looseness=-1

H-SPAR adopts a one-way coupling assumption: the environmental velocity field affects both Lagrangian particle transport and \gls{usv} motion, whereas disturbances induced by the vehicle are not fed back into the flow field or particle dynamics. Consequently, localized near-body effects, including wake generation, propeller wash, and vehicle-induced particle disturbance, are not resolved. This approximation is suitable for the mission-scale simulations considered here, where vehicle-induced flow is assumed to remain localized relative to the spatial extent of the environmental current field. Resolving these interactions would require two-way fluid--structure coupling or local \gls{cfd} around the vehicle and is left for future work.

\subsection{\textcolor{black}{Probabilistic Particle Sampling Model}}
\label{sec::sampling}

\textcolor{black}{To evaluate particle-monitoring missions, H-SPAR includes a mission-level probabilistic sampling model. The model does not resolve the detailed fluid-particle-pump interaction; instead, it estimates whether particles encountered by the \gls{usv} are sampled as a function of their relative motion with respect to the robot.}

\textcolor{black}{At each simulation time step, particles located within the sampling region of the \gls{usv} are treated as candidate particles. Let ${}^{r}\mathbf{x}_p$ denote the position of particle $p$ expressed in the robot frame, and let $r_s$ denote the sampling radius. A particle is considered inside the sampling region if $\|{}^{r}\mathbf{x}_p\| \leq r_s $. We denote this encounter event as $E_p: \|{}^{r}\mathbf{x}_p\| \leq r_s$. Given that particle $p$ is inside the sampling region, the probability of sampling the particle is modeled as a conditional probability:}
\begin{equation}
    P(S_p \mid E_p)
    =
    p_c(\mathbf{v}_{p/r}),
    \label{eq:conditional_sampling_probability}
\end{equation}
\textcolor{black}{where $S_p$ denotes the event that particle $p$ is sampled, $v_{p/r}$ is the relative speed between the particle and the robot, and $p_c$ is the particle capture probability function. The particle-robot relative velocity is defined as}
\begin{equation}
    \mathbf{v}_{p/r} = \mathbf{v}_p - \mathbf{v}_r,
    \label{eq:particle_robot_relative_velocity}
\end{equation}
\textcolor{black}{where $\mathbf{v}_p$ and $\mathbf{v}_r$ are expressed in the same reference frame.}

\textcolor{black}{For each candidate particle, a random variable $r \sim \mathcal{U}(0,1)$ is generated. The particle is considered sampled if $r<p_c(\mathbf{v}_{p/r})$}. \textcolor{black}{The capture probability is modeled as a Gaussian function centered at the design intake velocity of the sampling pump:}
\begin{equation} 
    p_c(\mathbf{v}_{p/r})
    =
    \exp\left(
    -\frac{\left(\|\mathbf{v}_{p/r}\|-v_0\right)^2}{2\sigma^2}
    \right),
    \label{eq:particle_capture_probability}
\end{equation}
where $\|\mathbf{v}_{p/r}\|$ is the magnitude of the fluid velocity relative to
the sampling nozzle, $v_0$ is the design intake velocity of the internal
sampling pump, and $\sigma$ controls the sensitivity of the capture
probability to deviations from the isokinetic condition. The maximum capture
probability occurs when $\|\mathbf{v}_{p/r}\|=v_0$, corresponding to isokinetic
sampling, in which the fluid approaches the nozzle at the same speed as it is drawn into the sampling system~\cite{weber2023underestimating}. 

Once a particle is sampled, it is removed from the particle set to prevent duplicate detections. This allows H-SPAR to compare monitoring strategies using sampling-oriented metrics, e.g. the number of encountered particles, the number of sampled particles, and the particle sampling rate.

%Equivalently, the overall sampling probability can be written as}
% \begin{equation}
%     P(S_p)=
%     \begin{cases}
%     p_c(\mathbf{v}_{p/r}), & \|{}^{r}\mathbf{x}_p\| \leq r_s,\\
%     0, & \text{otherwise}
%     \end{cases}.
%     \label{eq:overall_sampling_probability}
% \end{equation}

\section{Experiments}\label{experiments}
\textcolor{black}{H-SPAR supports the evaluation of energy-constrained \gls{usv} operations and pollutant-sampling strategies. This section presents two experiments designed to demonstrate these capabilities. First, current-aware mission planners are evaluated at both the planning and execution levels, and their performance is compared using the proposed metrics (Sec.~\ref{sec::exp1}). Second, H-SPAR is used to investigate how mission design influences particle-sampling performance in coverage-based monitoring tasks (Sec.~\ref{sec::exp2}).}

In all experiments, the \gls{usv} model is constrained by maximum linear and angular velocities of $8~\text{m/s}$ and $1.5~\text{rad/s}$, and maximum linear and angular accelerations of $2~\text{m/s}^2$ and $1~\text{rad/s}^2$, respectively. For local trajectory execution, we adopt the \gls{dwa}, a widely used method in mobile robotics~\cite{fox2002dynamic}. \gls{dwa} explicitly considers kinematic constraints while generating smooth and feasible trajectories. All experiments were conducted on a workstation equipped with a 13th-generation Intel Core i7-13700K processor with 16 physical cores and 24 threads, 16~GB of memory, and an NVIDIA GeForce RTX 4080 GPU.

\subsection{Experiment 1: Current-Aware \textcolor{black}{Planning and Execution}}
\label{sec::exp1}

\begin{figure}[!t]
%fig:planners
\centering

% ============================================================
% TOP ROW: PANEL (a)
% Centered independently in the column
% ============================================================
\makebox[\columnwidth][c]{%
\begin{tikzpicture}

% ============================================================
% (a) MEAN FLOW FIELD
% ============================================================
\begin{axis}[
    name=flow,
    width=0.4\columnwidth,
    height=0.3\columnwidth,
    axis equal image,
    xmin=-420,
    xmax=-80,
    ymin=100,
    ymax=340,
    xlabel={$X_{\mathrm{NED}}$ [m]},
    ylabel={$Y_{\mathrm{NED}}$ [m]},
    tick label style={font=\small},
    xlabel style={
        font=\small,
        yshift=6pt
    },
    ylabel style={
        font=\small,
        yshift=-5pt
    },
    colormap/viridis,
    point meta min=0,
    point meta max=6,
    colorbar,
    colorbar style={
        width=0.025\columnwidth,
        ylabel={Current speed [m/s]},
        ylabel style={font=\small},
        tick label style={font=\small}
    }
]

\addplot[
    quiver={
        u=\thisrow{Un},
        v=\thisrow{Vn},
        scale arrows=0.5
    },
    point meta=\thisrow{Speed},
    quiver/colored,
    -stealth,
    thin
]
table[
    x=X,
    y=Y,
    col sep=comma
] {csv_files/mean_flow_unique2.csv};

\end{axis}

% ------------------------------------------------------------
% Caption for (a)
% ------------------------------------------------------------
\node[
    anchor=north,
    font=\small
]
at ($(flow.outer south)+(0.5,-0cm)$)
{(a) Mean flow field};

\end{tikzpicture}%
}

% ============================================================
% SPACE BETWEEN ROWS
% ============================================================
\vspace{0.3cm}

% ============================================================
% BOTTOM ROW: PANELS (b) AND (c)
% Centered independently in the column
% ============================================================
\makebox[\columnwidth][c]{%
\begin{tikzpicture}

% ============================================================
% (b) PLANNED PATHS
% ============================================================
\begin{axis}[
    name=planned,
    at={(-0.225\columnwidth,0)},
    anchor=north,
    width=0.4\columnwidth,
    height=0.30\columnwidth,
    axis equal image,
    xmin=-420,
    xmax=-80,
    ymin=100,
    ymax=340,
    xlabel={$X_{\mathrm{NED}}$ [m]},
    ylabel={$Y_{\mathrm{NED}}$ [m]},
    tick label style={font=\small},
    xlabel style={
    font=\small,
    yshift=6pt
    },
    ylabel style={
    font=\small,
    yshift=-5pt
    },
]

% RRT*
\addplot[
    thick,
    solid,
    color=rrtColor,
    no marks
]
table[
    x=x,
    y=y,
    col sep=comma
] {csv_files/rrt/rrt_waypoints.csv};

% VF-RRT*
\addplot[
    thick,
    dashdotted,
    color=vfColor,
    no marks
]
table[
    x=x,
    y=y,
    col sep=comma
] {csv_files/vf-rrt/vf_waypoints.csv};

% SVF-RRT*
\addplot[
    thick,
    dashed,
    color=svfColor,
    no marks
]
table[
    x=x,
    y=y,
    col sep=comma
] {csv_files/svf-rrt/svf_waypoints.csv};

% Start
\addplot[
    only marks,
    mark=*,
    mark size=2pt,
    color=black
]
coordinates {(-400,200)};

% End
\addplot[
    only marks,
    mark=square*,
    mark size=2pt,
    color=black
]
coordinates {(-105,280)};

\end{axis}
\hspace{0.5cm}

% ============================================================
% (c) EXECUTED TRAJECTORIES
% ============================================================
\begin{axis}[
    name=executed,
    at={(0.225\columnwidth,0)},
    anchor=north,
    width=0.4\columnwidth,
    height=0.3\columnwidth,
    axis equal image,
    xmin=-420,
    xmax=-80,
    ymin=100,
    ymax=340,
    xlabel={$X_{\mathrm{NED}}$ [m]},
    ylabel={$Y_{\mathrm{NED}}$ [m]},
    tick label style={font=\small},
    xlabel style={
    font=\small,
    yshift=6pt
    },
    ylabel style={
        font=\small,
        yshift=-5pt
    },
]

% RRT*
\addplot[
    thick,
    solid,
    color=rrtColor,
    no marks,
    each nth point=4
]
table[
    x=x_ned,
    y=y_ned,
    col sep=comma
] {csv_files/rrt/trajectory.csv};

% VF-RRT*
\addplot[
    thick,
    dashdotted,
    color=vfColor,
    no marks,
    each nth point=8
]
table[
    x=x_ned,
    y=y_ned,
    col sep=comma
] {csv_files/vf-rrt/trajectory.csv};

% SVF-RRT*
\addplot[
    thick,
    dashed,
    color=svfColor,
    no marks,
    each nth point=4
]
table[
    x=x_ned,
    y=y_ned,
    col sep=comma
] {csv_files/svf-rrt/trajectory.csv};

% Start
\addplot[
    only marks,
    mark=*,
    mark size=2pt,
    color=black
]
coordinates {(-400,200)};

% End
\addplot[
    only marks,
    mark=square*,
    mark size=2pt,
    color=black
]
coordinates {(-110,280)};

\end{axis}

% ============================================================
% SUBFIGURE LABELS
% ============================================================
\node[
    anchor=north,
    font=\small
]
at ($(planned.outer south)+(0,0.0cm)$)
{(b) Planned paths};

\node[
    anchor=north,
    font=\small
]
at ($(executed.outer south)+(0,0cm)$)
{(c) Executed trajectories};

% ============================================================
% SHARED LEGEND: ALL ITEMS IN ONE LINE
% ============================================================
\coordinate (legendCenter) at
($(planned.south)!0.5!(executed.south)+(-0.2,-2cm)$);

% Legend box
\node[
    draw=gray!60,
    line width=0.3pt,
    rounded corners=1pt,
    minimum width=0.96\columnwidth,
    minimum height=0.38cm,
    inner sep=2pt
] at (legendCenter) {};

% RRT*
\draw[
    thick,
    solid,
    color=rrtColor
]
($(legendCenter)+(-0.45\columnwidth,0)$)
--
++(0.035\columnwidth,0);

\node[
    anchor=west,
    font=\scriptsize
]
at ($(legendCenter)+(-0.41\columnwidth,0)$)
{RRT*};

% VF-RRT*
\draw[
    thick,
    dashdotted,
    color=vfColor
]
($(legendCenter)+(-0.29\columnwidth,0)$)
--
++(0.035\columnwidth,0);

\node[
    anchor=west,
    font=\scriptsize
]
at ($(legendCenter)+(-0.25\columnwidth,0)$)
{VF-RRT*};

% SVF-RRT*
\draw[
    thick,
    dashed,
    color=svfColor
]
($(legendCenter)+(-0.065\columnwidth,0)$)
--
++(0.035\columnwidth,0);

\node[
    anchor=west,
    font=\scriptsize
]
at ($(legendCenter)+(-0.025\columnwidth,0)$)
{SVF-RRT*};

% Start
\node[
    circle,
    fill=black,
    inner sep=1.25pt
]
at ($(legendCenter)+(0.2\columnwidth,0)$)
{};

\node[
    anchor=west,
    font=\scriptsize
]
at ($(legendCenter)+(0.215\columnwidth,0)$)
{Start};

% End
\node[
    rectangle,
    fill=black,
    draw=black,
    minimum width=3.5pt,
    minimum height=3.5pt,
    inner sep=0pt
]
at ($(legendCenter)+(0.365\columnwidth,0)$)
{};

\node[
    anchor=west,
    font=\scriptsize
]
at ($(legendCenter)+(0.38\columnwidth,0)$)
{End};

\end{tikzpicture}%
}

% ============================================================
% MAIN CAPTION
% ============================================================
\caption{
Energy-aware operation in H-SPAR at the mission-planning and
execution-control levels.
(a) Average velocity vector field over all precomputed timestamps in
the \gls{usv} operating area.
(b) Planned paths generated by RRT*, VF-RRT*, and SVF-RRT*.
(c) Corresponding executed trajectories in H-SPAR.
}
\label{fig:exp1}
\end{figure}
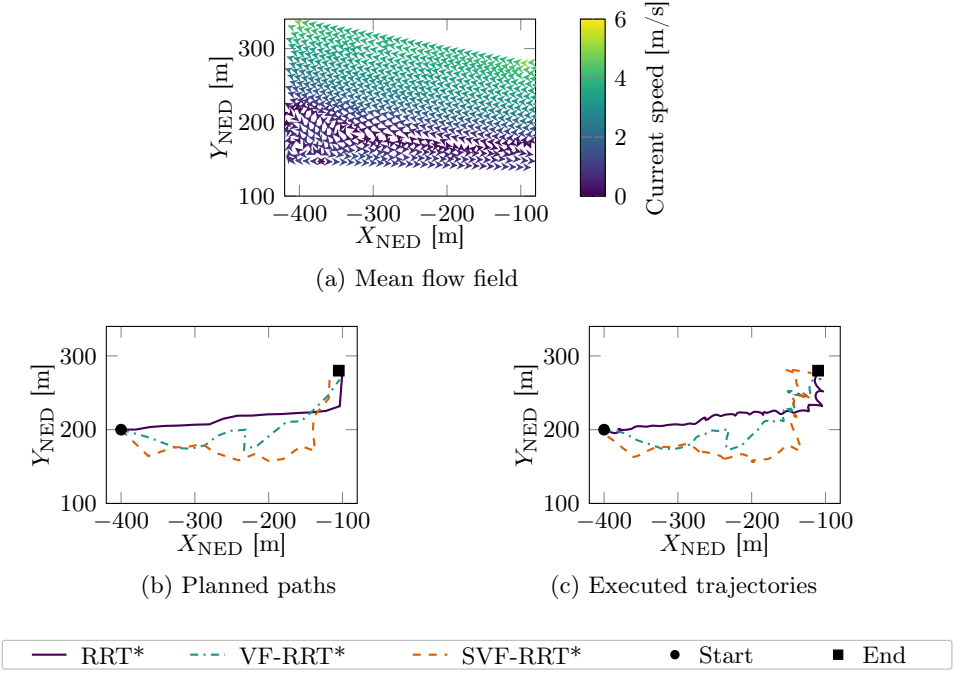

This experiment evaluates the paths generated by the current-aware planners and compares their predicted performance during planning with their realized performance during mission execution. Three planners are compared: the geometry-based RRT* baseline~\cite{rrt-star}, VF-RRT*~\cite{ko2013vf}, and SVF-RRT*~\cite{zhang2023svf}, as illustrated in Fig.~\ref{fig:exp1}. All planners generate a waypoint sequence using the same static costmap, while VF-RRT* and SVF-RRT* additionally incorporate the environmental velocity field into their planning costs. Since these planners do not explicitly consider temporal variations in the flow field, the time-averaged velocity field over all simulation timestamps is used during planning.

The waypoint sequence generated by each path planner is subsequently passed to the same \gls{dwa} motion planner for execution. The \gls{dwa} accounts for the kinematic constraints of the \gls{usv}, allowing the performance of the planned paths to be assessed under a common execution framework. At the planning level, path length and upstream cost are evaluated. At the execution level, the resulting trajectory length, mission time, path deviation, and executed upstream cost are reported.

As shown in Table~\ref{tab::plan_exec}, RRT* produces the shortest planned path but also incurs the highest planned upstream cost. This result demonstrates that minimizing geometric path length alone does not necessarily yield a favorable trajectory in the presence of environmental currents. In contrast, VF-RRT* and SVF-RRT* generate longer paths while substantially reducing the planned upstream cost, with SVF-RRT* achieving the lowest value among the evaluated planners.

\FloatBarrier
\begin{table}[!t]
    \color{black}
    %\captionsetup{width=\columnwidth}
\caption{Planning- and execution-level comparison of RRT*, VF-RRT*, and SVF-RRT* under time-averaged and temporally varying flow fields. Current-aware planners improve execution performance when operating under the same static flow field used for planning, but their advantage diminishes when the flow field varies over time.}
    \label{tab::plan_exec}
    \centering
    \small
    \renewcommand{\arraystretch}{0.9}
    \begin{tabular}{llccc}
        \hline
        Category & Metric & RRT* & VF-RRT* & SVF-RRT* \\
        \hline
        \multirow{2}{*}{Plan.}
        & Length (m) & \textbf{348.2} & 358.4 & 408.2 \\
        & Upstream cost & 1238.98 & 667.71 & \textbf{379.49} \\
        \hline
        \multirow{4}{*}{\shortstack[l]{Exec.\\(Temporal Field)}}
        & Traj. length (m) & \textbf{392.946} & 483.7 & 546.7 \\
        & Time (s) & \textbf{1127.0} & 1651.5 & 1349.1 \\
        & Deviation (m) & \textbf{2.82} & 5.6 & 6.7 \\
        & Upstream cost & 1312.8 & 1054.2 & \textbf{765.2} \\
        \hline
        \multirow{4}{*}{\shortstack[l]{Exec.\\(Average Field)}}
        & Traj. length (m) & \textbf{389.53} & 394.64 & 446.08 \\
        & Time (s) & 1230.97 & 1025.97 & \textbf{993.96} \\
        & Deviation (m) & 1.95 & \textbf{1.93} & 2.47 \\
        & Upstream cost & 1401.27 & 789.55 & \textbf{486.55} \\
        \hline
        % \multirow{4}{*}{\shortstack[l]{Exec.\\(Constant)}}
        % & Traj. length (m) & \textbf{386.788} & 492.02 & 484.48 \\
        % & Time (s) & \textbf{1031.96} & 1541.07 & 1583.72 \\
        % & Deviation (m) & \textbf{1.57} & 3.79 & 4.99 \\
        % & Upstream cost & 1365.73 & 835.512 & \textbf{544.43} \\
        % \hline
    \end{tabular}
\end{table}

When the planned trajectories are executed under the same time-averaged velocity field used during planning, the current-aware planners provide lower mission times and executed upstream costs while maintaining trajectory lengths and path deviations comparable to those of RRT*. This indicates that incorporating current information into the path planning objective can improve execution performance when the environmental model used for planning is representative of the conditions encountered by the vehicle.

The performance changes, however, when the \gls{usv} is exposed to the original time-varying velocity field during execution. Under these conditions, RRT* exhibits a shorter executed trajectory and lower path deviation than the current-aware planners. This degradation highlights a limitation of VF-RRT* and SVF-RRT*: because their planning costs are computed using a time-averaged representation of the current field, the resulting paths may become less favorable when the instantaneous flow differs from the field assumed during planning.

A second factor contributing to this behavior is the mismatch between path optimization and vehicle-level feasibility. The path planners do not explicitly account for the non-holonomic motion constraints of the \gls{usv} or the hydrodynamic forces experienced during execution, leaving the motion planner responsible for compensating for these effects. In the considered scenario, RRT* selects a comparatively direct route that largely opposes the dominant current while avoiding frequent lateral crossings of the flow. In contrast, VF-RRT* and SVF-RRT* favor regions with lower upstream cost and may introduce sharper transitions across the current field, as shown in Fig.~\ref{fig:exp1}. Although these paths are favorable according to the planning-level current-aware objective, they can require larger corrective actions during execution, particularly when the flow varies temporally. This results in increased path deviation and mission time in the time-varying-current case, as reported in Table~\ref{tab::plan_exec}.

Overall, these results demonstrate the value of evaluating current-aware planning not only according to planning-level cost metrics but also under closed-loop execution. They further indicate that current-aware path planners based on static or time-averaged flow representations can provide substantial benefits when the assumed flow field remains representative, while their advantage may decrease when temporal variations in the environment are not incorporated into the planning process.

% \begin{figure*}[t]
%     \centering

%     \begin{subfigure}[h]{0.3\textwidth}
%         \centering
%         \includesvg[height=5cm, keepaspectratio]{figures/0.svg}
%         \caption{\textit{Heading angle=0}}
%         \label{fig:velocity_field}
%     \end{subfigure}
%     \hfill
%     \begin{subfigure}[h]{0.3\textwidth}
%         \centering
%         \includesvg[height=5cm, keepaspectratio]{figures/45.svg}
%         \caption{\textit{Heading angle=45}}
%         \label{fig:planning_paths}
%     \end{subfigure}
%     \hfill
%     \begin{subfigure}[h]{0.3\textwidth}
%         \centering
%         \includesvg[height=6cm, keepaspectratio]{figures/90.svg}
%         \caption{\textit{Heading angle=90}}
%         \label{fig:execution_trajectories}
%     \end{subfigure}

%     \caption{Comparison of sweeping patterns over a region of interest using three different heading angles. Detected particles are visualized as heatmaps in each plot.}
%     \label{fig:heatmap_p}
% \end{figure*}

% Preamble required: \usepackage{booktabs}

\subsection{Experiment 2: Coverage-Based Particle Monitoring}
\label{sec::exp2}
This experiment evaluates coverage-based particle monitoring under hydrodynamic conditions, with emphasis on sampling performance rather than solely on \gls{usv} motion. In current-influenced environments, coverage planning requires several mission-level decisions, including the extent of the coverage region, sweep orientation, line spacing, and vehicle heading configuration. These choices cannot be determined from geometry alone because the surrounding flow affects mission duration, current-related cost, particle transport, and the likelihood of particle encounters during execution.

In this experiment, we have tested the effect of sweep pattern on sampling performance. We employ boustrophedon sweep-based strategy to generate waypoints for systematic area monitoring~\cite{sudha2024coverage}. In H-SPAR, boustrophedon waypoints are automatically generated within a user-defined region of interest using the method described by Zhao~et~al.~\cite{zhao2024energy}. Different sweep orientations are then executed under the prescribed velocity field and particle distribution to evaluate their effect on monitoring and sampling performance (Fig.~\ref{fig:heatmap_p}).

To maintain a continuous and comparable particle influx throughout the experiment, 100 particles are released from the upstream boundary every 10~s. Particle trajectories are simulated under both a uniform flow field and the temporally varying flow fields provided by H-SPAR (Fig.~\ref{fig:heatmap_p}a). This enables the influence of hydrodynamic particle transport on the resulting spatial particle distribution and sampling opportunities to be evaluated (Fig.~\ref{fig:heatmap_p}b-d).

\subsubsection{Ablation Study}
\label{sec:coverage-ablation}
To isolate the contribution of these modeling components, we conduct a sequence of ablation experiments in which temporal flow variation, hydrodynamic wrench forces, and probabilistic sampling are introduced incrementally, as summarized in Table~\ref{tab:experimental_results}. For probabilistic sampling, the intake design velocity is set to $0.5~m/s$, with a standard deviation of $0.2$, according to Eq.~\eqref{eq:particle_capture_probability}.
\begin{table}[b]
\color{black}
\centering
\caption{Particle sampling performance across sweep orientations and modeling configurations. Sampling rate is used to normalize the effect of mission duration on particle-encounter opportunities. Hydrodynamic wrench forces do not change the sweep orientation with the highest sampling rate, whereas the use of probabilistic sampling and spatially varying flow fields can change the preferred sweep orientation.}

\label{tab:experimental_results}
\footnotesize
\begin{tabular}{lcccccc}
\toprule
& & & & \multicolumn{3}{c}{\textbf{Particle Count (Rate)}} \\
\cmidrule(lr){5-7}
\textbf{Config} & \textbf{Flow Field} & \textbf{Sampling} & \textbf{Wrench} & $0^\circ$ & $45^\circ$ & $90^\circ$ \\
\midrule
\textbf{1} & Uniform & Det. & \nomark & 373 (0.43) & \textbf{463} (0.45)  & 286 (\textbf{0.46}) \\
\textbf{2} & Uniform & Det. & \checkmark & 469 (0.44) & \textbf{489} (0.36) & 346 (\textbf{0.46}) \\
\textbf{3} & Uniform & Prob. & \checkmark & 122 (\textbf{0.12}) & \textbf{130} (0.10) & 68 (0.09)  \\
\textbf{4} & H-SPAR & Det. & \nomark & 288 (0.39) & \textbf{383} (\textbf{0.42}) & 258 (0.41) \\
\textbf{5} & H-SPAR & Det. & \checkmark & 316 (0.38) & \textbf{438} (\textbf{0.41}) & 279 (0.37) \\
% \textbf{5} & H-SPAR & Prob. & \xmark & 90 (0.12) & 100 (0.1) & 62 (0.08) \\
\textbf{6} & H-SPAR & Prob. & \checkmark & 101 (\textbf{0.11}) & \textbf{114} (0.09) & 86 (0.10)\\
\bottomrule
\end{tabular}
\end{table}

\definecolor{barZero}{HTML}{4C78A8}
\definecolor{barFortyFive}{HTML}{F58518}
\definecolor{barNinety}{HTML}{9E9E9E}

\definecolor{barZero}{HTML}{4C78A8}
\definecolor{barFortyFive}{HTML}{F58518}
\definecolor{barNinety}{HTML}{9E9E9E}

\begin{figure}[p]
\centering

% ============================================================
% ROW 1: (a), (b), (c)
% ============================================================

% -------------------- (a) Sweep Region -----------------------
\begin{subfigure}[b]{0.47\textwidth}
    \centering
    \includegraphics[
        height=3cm,
        width=\textwidth,
        keepaspectratio
    ]{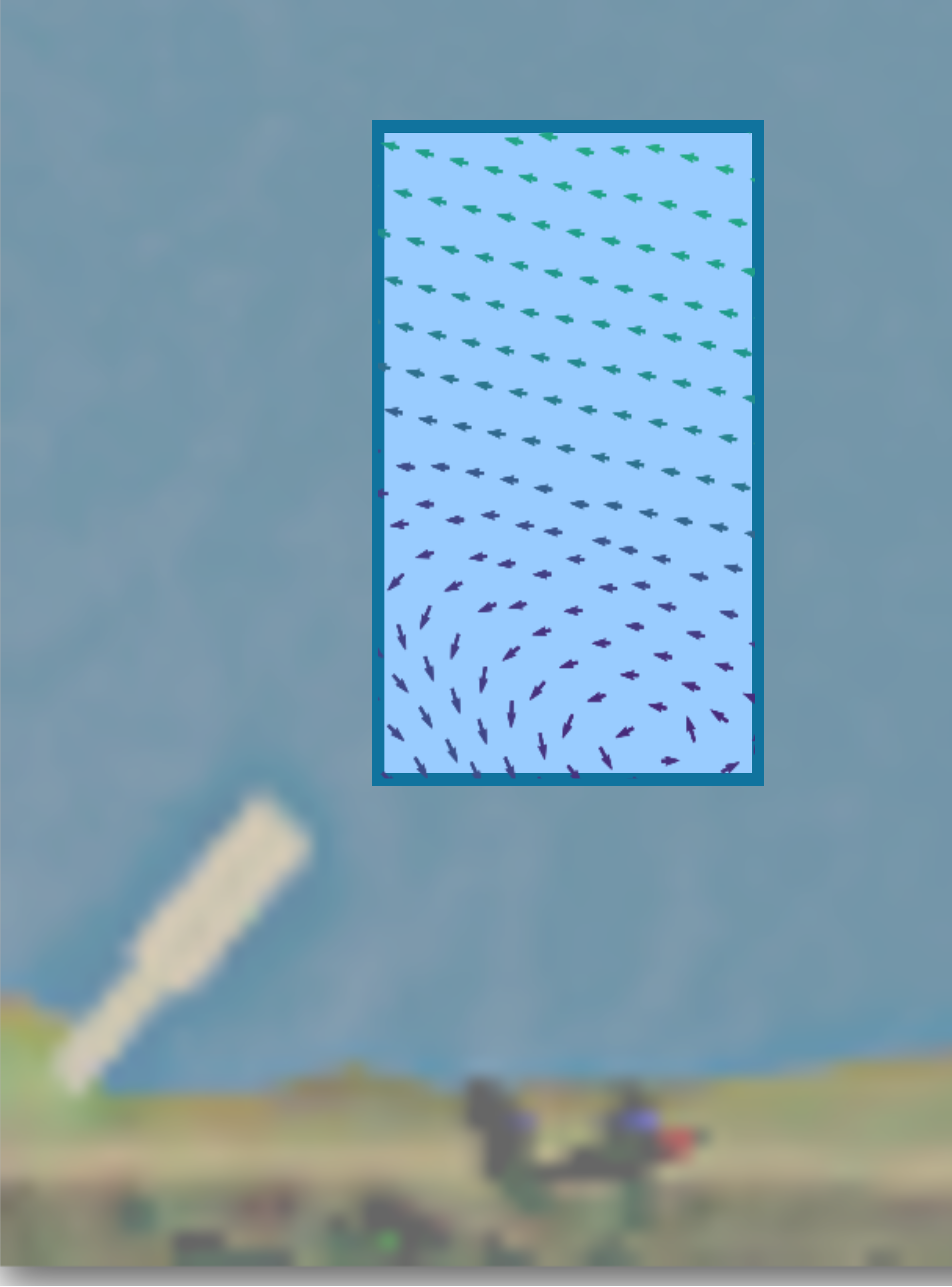}
    \caption{Sweep Region}
    \label{fig:placeholder}
\end{subfigure}\hfill
%
% -------------------- (b) 0 degree ---------------------------
\begin{subfigure}[b]{0.52\textwidth}
    \centering
    \begin{tikzpicture}
    \begin{axis}[
        scale only axis,
        width=2.1cm,
        height=2.2cm,
        enlargelimits=false,
        xlabel={X (NED) [m]},
        ylabel={Y (NED) [m]},
        xlabel style={font=\small, yshift=2pt},
        ylabel style={
            at={(axis description cs:-0.35,0.5)},
            anchor=south,
            font=\small
        },
        tick label style={font=\small},
        xtick={-440,-420},
        ytick={200,220,240},
        grid=major,
        grid style={dashed,gray!30},
        colorbar,
        colorbar style={
            title={Time [s]},
            title style={font=\small,yshift=-2pt},
            tick label style={font=\small},
            width=0.07cm,
            height=2.0cm,
            ytick distance=200,
            xshift=-5pt,
            yshift=-4pt
        },
        colormap/viridis,
        xmin=-445,
        xmax=-415,
        ymin=195,
        ymax=245
    ]

    \addplot[
        color=gray!60,
        line width=0.8pt,
        no markers
    ]
    table[
        x=x_ned,
        y=y_ned,
        col sep=comma
    ]{csv_files/coverage_csvs/0_traj.csv};

    \addplot[
        scatter,
        only marks,
        scatter src=explicit,
        mark=*,
        mark size=0.8pt,
        draw=black!80,
        point meta=explicit
    ]
    table[
        x=robot_x,
        y=robot_y,
        meta=sim_time,
        col sep=comma
    ]{csv_files/coverage_csvs/0_detections.csv};

    \end{axis}
    \end{tikzpicture}
    \caption{$0^\circ$ Sweep}
\end{subfigure}\par\medskip

% -------------------- (c) 45 degree --------------------------
\begin{subfigure}[b]{0.47\textwidth}
    \centering
    \begin{tikzpicture}
    \begin{axis}[
        scale only axis,
        width=2.1cm,
        height=2.2cm,
        enlargelimits=false,
        xlabel={X (NED) [m]},
        ylabel={Y (NED) [m]},
        xlabel style={font=\small,yshift=2pt},
        ylabel style={
            at={(axis description cs:-0.35,0.5)},
            anchor=south,
            font=\small
        },
        tick label style={font=\small},
        xtick={-440,-420},
        ytick={200,220,240},
        grid=major,
        grid style={dashed,gray!30},
        colorbar,
        colorbar style={
            title={Time [s]},
            title style={font=\small,yshift=-2pt},
            tick label style={font=\small},
            width=0.07cm,
            height=2.0cm,
            ytick distance=200,
            xshift=-5pt,
            yshift=-4pt
        },
        colormap/viridis,
        xmin=-445,
        xmax=-415,
        ymin=195,
        ymax=245,
        legend columns=3,
        legend to name=shared_legend,
        legend style={
            font=\scriptsize,
            draw=gray!30,
            fill=white,
            inner sep=4pt,
            /tikz/every even column/.append style={
                column sep=0.5cm
            }
        }
    ]

    \addplot[
        color=gray!60,
        line width=0.8pt,
        no markers
    ]
    table[
        x=x_ned,
        y=y_ned,
        col sep=comma
    ]{csv_files/coverage_csvs/45_traj.csv};
    \addlegendentry{USV trajectory}

    \addplot[
        scatter,
        only marks,
        scatter src=explicit,
        mark=*,
        mark size=0.8pt,
        draw=black!80,
        point meta=explicit
    ]
    table[
        x=robot_x,
        y=robot_y,
        meta=sim_time,
        col sep=comma
    ]{csv_files/coverage_csvs/45_detections.csv};
    \addlegendentry{Detected Particles}

    \addlegendimage{
        area legend,
        fill=barZero!85,
        draw=barZero,
        line width=0.5pt
    }
    \addlegendentry{\(0^\circ\) Sweep}

    \addlegendimage{
        area legend,
        fill=barFortyFive!85,
        draw=barFortyFive,
        line width=0.5pt
    }
    \addlegendentry{\(45^\circ\) Sweep}

    \addlegendimage{
        area legend,
        fill=barNinety!85,
        draw=barNinety,
        line width=0.5pt
    }
    \addlegendentry{\(90^\circ\) Sweep}

    \end{axis}
    \end{tikzpicture}
    \caption{$45^\circ$ Sweep}
\end{subfigure}
\begin{subfigure}[b]{0.47\textwidth}
    \centering
    \begin{tikzpicture}
    \begin{axis}[
        scale only axis,
        width=2.1cm,
        height=2.2cm,
        enlargelimits=false,
        xlabel={X (NED) [m]},
        ylabel={Y (NED) [m]},
        xlabel style={font=\small,yshift=2pt},
        ylabel style={
            at={(axis description cs:-0.35,0.5)},
            anchor=south,
            font=\small
        },
        tick label style={font=\small},
        xtick={-440,-420},
        ytick={200,220,240},
        grid=major,
        grid style={dashed,gray!30},
        colorbar,
        colorbar style={
            title={Time [s]},
            title style={font=\small,yshift=-2pt},
            tick label style={font=\small},
            width=0.07cm,
            height=2.0cm,
            ytick distance=200,
            xshift=-5pt,
            yshift=-4pt
        },
        colormap/viridis,
        xmin=-445,
        xmax=-415,
        ymin=195,
        ymax=245
    ]

    \addplot[
        color=gray!60,
        line width=0.8pt,
        no markers
    ]
    table[
        x=x_ned,
        y=y_ned,
        col sep=comma
    ]{csv_files/coverage_csvs/90_traj.csv};

    \addplot[
        scatter,
        only marks,
        scatter src=explicit,
        mark=*,
        mark size=0.8pt,
        draw=black!80,
        point meta=explicit
    ]
    table[
        x=robot_x,
        y=robot_y,
        meta=sim_time,
        col sep=comma
    ]{csv_files/coverage_csvs/90_detections.csv};

    \end{axis}
    \end{tikzpicture}
    \caption{$90^\circ$ Sweep}
\end{subfigure}\par\medskip

\begin{subfigure}[b]{0.6\textwidth}
    \centering
    \begin{tikzpicture}
    \begin{axis}[
        scale only axis,
        width=4cm,
        height=2cm,
        ybar=2.5pt,
        bar width=5pt,
        enlarge x limits=0.20,
        ymin=0,
        ymax=1250,
        ytick={0,500,1000},
        ylabel={Value},
        ylabel style={
            at={(axis description cs:-0.20,0.4)},
            anchor=south,
            font=\small
        },
        symbolic x coords={
            Mission Time,
            Upstream Cost,
            Particle Count
        },
        xtick=data,
        tick label style={font=\small},
        xticklabel style={
            rotate=12,
            anchor=north east,
            font=\small
        },
        ymajorgrids=true,
        grid style={dashed,gray!30},
        nodes near coords,
        nodes near coords style={
            font=\scriptsize,
            color=black!85,
            /pgf/number format/fixed,
            /pgf/number format/precision=0,
            rotate=90,
            anchor=west,
            xshift=0pt
        }
    ]

    \addplot[
        fill=barZero!85,
        draw=barZero,
        line width=0.5pt
    ] coordinates {
        (Mission Time,761)
        (Upstream Cost,114)
        (Particle Count,288)
    };

    \addplot[
        fill=barFortyFive!85,
        draw=barFortyFive,
        line width=0.5pt
    ] coordinates {
        (Mission Time,811)
        (Upstream Cost,135)
        (Particle Count,383)
    };

    \addplot[
        fill=barNinety!85,
        draw=barNinety,
        line width=0.5pt
    ] coordinates {
        (Mission Time,573)
        (Upstream Cost,204)
        (Particle Count,258)
    };

    \end{axis}
    \end{tikzpicture}
    \caption{Metrics}
\end{subfigure}

% ============================================================
% SHARED LEGEND
% ============================================================
\vspace{0.25cm}

\par\noindent
\makebox[\textwidth][c]{%
    \pgfplotslegendfromname{shared_legend}%
}

% ============================================================
% CAPTION
% ============================================================
\caption{\textcolor{black}{
Comparison of coverage patterns over a region of interest using three
sweep orientations, assuming deterministic sampling and neglecting
wrench effects on the \gls{usv}.
(a)~Region of interest used to generate the coverage paths with vector
field visualized inside the region.
\gls{usv} trajectories are generated within the region using
(b)~$0^\circ$, (c)~$45^\circ$, and (d)~$90^\circ$ sweep orientations.
(e)~Comparison of the corresponding mission metrics shows that the
$45^\circ$ sweep pattern produces the highest number of particle
encounters, primarily because of its longer mission duration.
However, the highest detection rate is achieved with the $0^\circ$
sweep, which also exhibits the lowest upstream cost. These results
indicate that the optimal sweep direction depends on the combined
effects of mission duration, the spatiotemporal flow field, particle
transport, and the relative velocity between the \gls{usv} and the
particles.
}}
\label{fig:heatmap_p}

\end{figure}
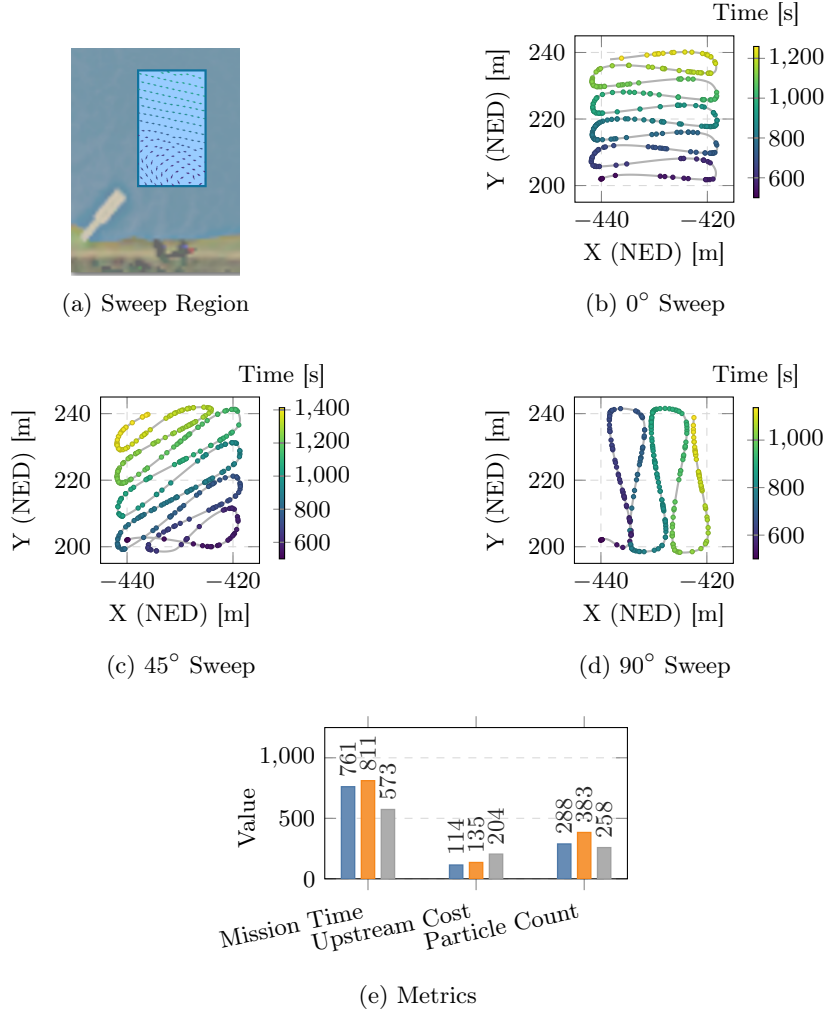

The number of sampled particles generally increases with mission duration. Consequently, the $45^\circ$ sweep, which has the longest mission time in Fig.~\ref{fig:heatmap_p}e, also produces the highest absolute particle count across all configurations in Table~\ref{tab:experimental_results}. Absolute particle count alone, however, is not sufficient for comparing coverage strategies because longer missions provide more opportunities for particle encounters. The normalized sampling rate therefore provides a more informative measure of sweep efficiency.

As shown in Table~\ref{tab:experimental_results}, the sweep orientation associated with the highest sampling rate depends on the environmental and sampling assumptions. Under a uniform flow field with deterministic sampling, the $90^\circ$ sweep achieves the highest sampling rate, indicating that trajectories oriented perpendicular to the dominant flow can increase particle encounters. When probabilistic sampling is introduced, however, the $0^\circ$ sweep achieves the highest rate. This change results from the dependence of the capture probability on the relative velocity between the \gls{usv} and the particles, as described by Eq.~\eqref{eq:particle_capture_probability}.

The relationship becomes more complex under the spatiotemporally varying H-SPAR flow field. With deterministic sampling, the $45^\circ$ sweep produces the highest sampling rate, suggesting that its trajectory provides favorable encounters across the spatially varying current field. When probabilistic sampling is enabled, the $0^\circ$ sweep instead achieves the highest rate. This result indicates that maximizing the number of geometric particle encounters does not necessarily maximize the number of successfully sampled particles when capture depends on the relative particle-vehicle velocity.

Introducing hydrodynamic wrench forces changes the resulting particle counts and sampling rates but does not consistently change the sweep orientation associated with the highest rate. This suggests that, provided the \gls{usv} remains capable of tracking the prescribed coverage waypoints, particle transport and the relative velocity governing the sampling process can have a stronger influence on sampling performance than trajectory perturbations caused by the hydrodynamic wrench.

Overall, this experiment demonstrates that the preferred coverage orientation cannot be selected from geometric coverage considerations alone. The optimal sweep direction depends on the interaction among mission duration, the spatiotemporal flow field, particle transport, and the relative velocity between the \gls{usv} and particles. These results illustrate the value of jointly modeling hydrodynamics, vehicle execution, particle transport, and probabilistic sampling when evaluating marine particle-monitoring strategies.

{\color{black}\section{Conclusion}\label{conclusion}

This paper presented H-SPAR, a hydrodynamic-aware simulation framework that integrates flow-field modeling, Lagrangian particle transport, and \gls{ros2}/Gazebo-based \gls{usv} autonomy using a shared environmental representation. Experiments demonstrated that the shortest geometric path is not necessarily the most effective under spatially varying currents, while flow-aware planners can reduce upstream motion and improve mission performance at the cost of potential tracking deviations. Coverage experiments further showed that the preferred sweep orientation depends on mission duration, hydrodynamic cost, particle transport, and the sampling model, particularly when capture probability is governed by particle--vehicle relative velocity. These findings highlight the importance of evaluating marine monitoring strategies using planning-, execution-, and sampling-level metrics rather than geometric measures or total particle counts alone. Future work will validate simulated particle transport and vehicle drift against physical data, extend H-SPAR to three-dimensional and dynamically updated flow fields, incorporate higher-fidelity sensing and capture models, and evaluate additional rule-based and learning-based planning strategies.}

\subsection*{Data availability}

The H-SPAR source code and simulation data supporting
the findings of this study are publicly available on
Zenodo at \url{https://doi.org/10.5281/zenodo.23068670}.
The development repository is available at
\url{https://github.com/naviiidz/h-spar-sim}.

\bibliography{sn-bibliography}% common bib file
\end{document}